\pdfoutput=1

\documentclass[11pt]{article}

\usepackage{ACL2023}

\usepackage{times}
\usepackage{latexsym}

\usepackage[T1]{fontenc}

\usepackage[utf8]{inputenc}

\usepackage{microtype}

\usepackage{inconsolata}

\usepackage{graphicx}
\usepackage{enumerate}
\usepackage{booktabs}

\usepackage{color}
\usepackage{tcolorbox}
\usepackage{CJK}
\usepackage{adjustbox}
\tcbset{width=0.9\textwidth,boxrule=0pt,colback=red,arc=0pt,auto outer arc,left=0pt,right=0pt,boxsep=5pt}

\usepackage{float}
\usepackage{subfigure}

\usepackage[english]{babel}
\usepackage{blindtext}

\newcommand*{\affaddr}[1]{#1} 
\newcommand*{\affmark}[1][*]{\textsuperscript{#1}}
\newcommand*{\email}[1]{\texttt{#1}}

\title{A Memory Model for Question Answering from Streaming Data Supported by Rehearsal and Anticipation of Coreference Information}

\author{%
Vladimir Araujo\affmark[1,2], Alvaro Soto\affmark[2], Marie-Francine Moens\affmark[1]\\
\affaddr{\affmark[1]KU Leuven},
\affaddr{\affmark[2]Pontificia Universidad Católica de Chile}\\
\email{vgaraujo@uc.cl},
\email{asoto@ing.puc.cl},
\email{sien.moens@kuleuven.be}
}

\begin{document}
\maketitle
\begin{abstract}
Existing question answering methods often assume that the input content (e.g., documents or videos) is always accessible to solve the task. 
Alternatively, memory networks were introduced to mimic the human process of incremental comprehension and compression of the information in a fixed-capacity memory. 
However, these models only learn how to maintain memory by backpropagating errors in the answers through the entire network. 
Instead, it has been suggested that humans have effective mechanisms to boost their memorization capacities, such as rehearsal and anticipation. 
Drawing inspiration from these, we propose a memory model that performs rehearsal and anticipation while processing inputs to memorize important information for solving question answering tasks from streaming data.
The proposed mechanisms are applied self-supervised during training through masked modeling tasks focused on coreference information.
We validate our model on a short-sequence (bAbI) dataset as well as large-sequence textual (NarrativeQA) and video (ActivityNet-QA) question answering datasets, where it achieves substantial improvements over previous memory network approaches. 
Furthermore, our ablation study confirms the proposed mechanisms' importance for memory models.
\end{abstract}

\section{Introduction}



The question answering (QA) task is one of the most important and challenging tasks in natural language processing (NLP). A significant advance has been seen in this subject thanks to models based on deep learning \cite{10.5555/2969239.2969428,seo2017bidirectional,chen-etal-2017-reading,devlin-etal-2019-bert}. However, these models assume that the whole input (e.g., sentences, paragraphs, etc.) can always be accessed while answering the question. This is a reasonable and practical approach if the input sequence is short but becomes less effective and efficient as the input length grows. These models are also cognitively implausible, lacking incremental human language processing \cite{Tanenhaus1995,keller-2010-cognitively} (or visual \cite{Goldstone1998}), which could enable the models for online tasks where data input is a stream, such as discourse, dialog or video processing.

\begin{figure}
  \centering
  \includegraphics[width=0.48\textwidth]{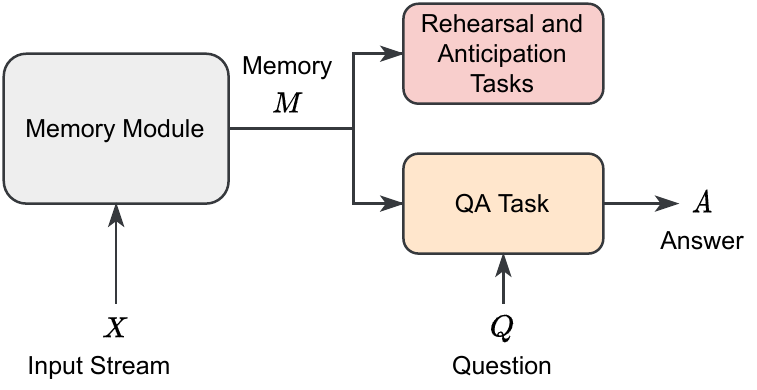}
  \caption{A rehearsal and anticipatory memory model (RAM) for question answering. The model is fed with a stream of data $X$, incrementally processed to fill a memory $M$. Such memory is used to obtain question-related $Q$ clues to provide an answer $A$. Pretext self-supervised tasks improve memorization by continually rehearsing and anticipating coreference information.}
  \label{fig:overview}
\end{figure}

Memory-augmented neural networks (MANNs) \cite{arxiv.1410.5401} offer a solution to this problem as they introduce mechanisms to compress and remember the input contents. MANNs have shown effectiveness in various tasks \cite{Graves2016,liu-etal-2019-referential,pmlr-v119-le20b}, including question answering \cite{miller-etal-2016-key,10.5555/3045390.3045643,han-etal-2019-episodic,pmlr-v139-zhang21ac}. In the latter, given a data sequence, the models process the input incrementally, capturing relevant information in the memory to answer a given query.
This is akin to the daily behavior of human beings. For instance, our mental state is updated when we encounter a new text segment while reading \cite{Traxler1997}. Later, we can use the memorized information to solve a specific task \cite{Moscovitch2016}.

Despite the success of MANNs and their similarity to human memory processes, they still have one significant limitation. Most of these models rely on learning a single task to maintain their memory; for example, a MANN for QA must learn what to store in memory while learning to answer questions. Instead, some studies have suggested that human beings are endowed with specific mechanisms for memorization. On the one hand, rehearsal \cite{Waugh1965,Craik1973} is a mechanism humans use to memorize information more effectively by repeating information over and over to be remembered. On the other hand, anticipation \cite{Hawkins2004-lp,Wittmann2007,Cole2015} suggests that our memory could lead to predictions of upcoming material. Furthermore, both processes are potentially guided by coreference information \cite{jaffe-etal-2020-coreference}, an integral part of discourse comprehension.


In this work, we propose a \textbf{R}ehearsal and \textbf{A}nticipatory \textbf{M}emory model (RAM) consisting of a memory module that uses a fusion module to integrate incoming information from a data stream to solve the QA task. This model is supported by two pretext tasks respectively inspired by human rehearsal and anticipation processes to enhance memory representation. These tasks are based on masked modeling \cite{devlin-etal-2019-bert} of coreference information, allowing the model to anticipate a coreferent and link it to the past through memory and rehearsal.
We validate our proposed model with datasets of short synthetic text sequences, long realistic text sequences, and video question answering. Results show that our model significantly outperforms previous well-known memory approaches due to the enhanced memory representation achieved by using the pretext tasks.




\section{Proposed Method}

In this section, we introduce our model, which processes inputs and builds memory representations incrementally from a data stream, used later to provide an answer to a question (Figure~\ref{fig:overview}).
RAM leverages attention and gating mechanisms along with masked modeling-based self-supervision to create a simple but effective method to improve memory representation and memorization.
We first describe the problem formulation of QA with MANNs. We then present our basic memory architecture for encoding input and decoding an answer. Finally, we introduce our novel self-supervised mechanisms inspired by rehearsal and anticipation guided by coreference information.

\subsection{Problem Formulation}
Let $X$ be an input stream and $Q$ a question related to the content of $X$. Standard QA approaches consist of a model $G(X,Q)$ that is trained to predict an answer $A$. In an incremental processing setup \cite{han-etal-2019-episodic}, the model does not have access to the whole input, so a fixed-size memory $M$ is used to compress the input stream $X$ one step at a time. Then a model $G(M,Q)$ learns to infer the answer $A$ for any relevant $Q$.

\begin{figure}
  \centering
  \includegraphics[width=0.48\textwidth]{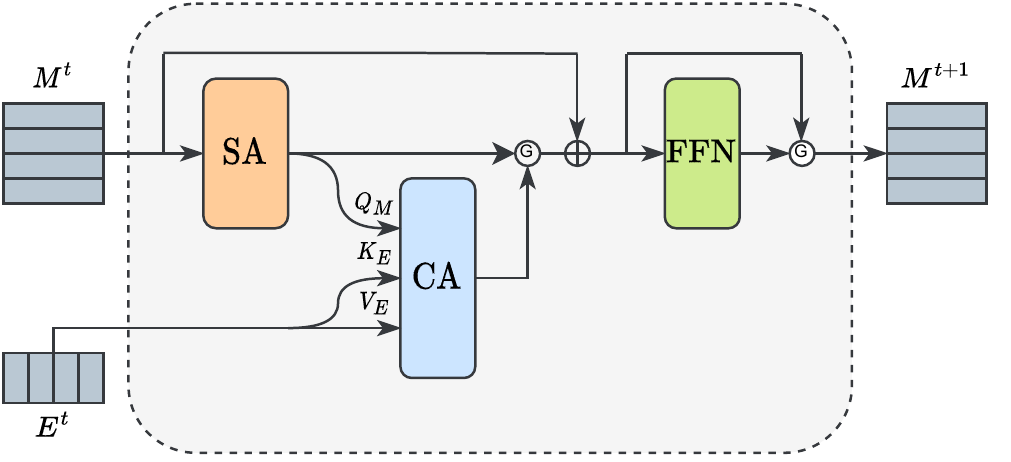}
  \caption{Our memory module consists of a self-attention ($\mathrm{SA}$) and cross-attention ($\mathrm{CA}$) layer that fuses the information from input $E^t$ to memory $M^t$ at step $t$. It also uses a gating mechanism to allow it to forget.}
  \label{fig:memory}
\end{figure}

\subsection{Memory Model Overview}

Here we present our basic model that incrementally processes the inputs and integrates the relevant information into memory to then answer a question.

\paragraph{Input Encoding:}

The input to our model is a sequence of segments (sequence of sentences or images in our experiments). A segment $X^t=\{ x^t_n \}^N_{n=1}$ at step $t$ with $N$ tokens is encoded by projecting each token into the latent space by using a linear function $F$. To inject information about the position of the tokens in the segment, we add learnable position embeddings $PE$. In this way, we obtain the final token embeddings of the segment $E^t=\{ e^t_n \}^N_{n=1}$ with dimension $d_{model}$. We use this procedure to encode the question as well.

\begin{equation}\label{eq:encoding}
E^t = F(X^t) + PE
\end{equation} 

Note that this form of representation allows the model to granularly capture and memorize the relevant elements of each segment, which differs from previous approaches that rely on a sequence of sum-pooled representations \cite{10.5555/3495724.3497539,pmlr-v119-le20b}.

\paragraph{Memory Module:}
Our model is augmented with a parametric memory $M=\{ m_k \}_{k=1}^K$ with $K$ slots intended to compress the information from the input stream effectively.
For this we implement a module that borrows the information fusion idea from \cite{fiber2022} and extends it to integrate information into the memory incrementally.

As shown in Figure~\ref{fig:memory}, at step $t$, our module receives two inputs: a segment representation $E^t$ and a memory $M^t$ to be updated. First, a self-attention layer ($\mathrm{SA}$) is used over the memory to allow the slots to interact with each other. Then, a cross-attention (CA) operation is performed to merge the tokens' information into memory (Eq.~\ref{eq:ca}). Note that the query matrix $Q_M$ (with dimension $d_{k}$) is computed from the memory and the key $K_E$ and value $V_E$ matrices from the segment. We employ a residual connection around the two layers.

As a result, we obtain an intermedial memory
state $\hat{M^t}$ (Eq.~\ref{eq:m_}) with $K$ slots, which has been updated with information from step $t$. 
Intuitively each slot can, for example, describe an object or an entity in the input or composition of them.


\begin{equation}\label{eq:ca}
\mathrm{CA}(M, E) = softmax\left( \frac{Q_M K_M^\top}{\sqrt{d_k}}\right) V_E
\end{equation}

\begin{equation}\label{eq:m_}
 \hat{M^t} = \mathrm{CA}( \mathrm{SA}(M^t), E^t)
\end{equation}

To allow the model to “forget,” an essential ability for memory networks, we employ a gating mechanism proposed by \citet{hutchins2022blockrecurrent}. 
First, the memory candidates are computed (Eq.~\ref{eq:gating1}). Unlike the standard LSTM gating, the candidate only is computed using the intermediate memory state $\hat{M^t}$ as it indirectly depends on $M^t$. Later, the input (Eq.~\ref{eq:gating2}) and forget gates (Eq.~\ref{eq:gating3}) which are used to obtain the new memory state (Eq.~\ref{eq:gating4}) are calculated. Note that the matrices $W$ and bias vectors $b$ are trainable.

\begin{equation}\label{eq:gating1}
z_t^k = tanh(W_z \hat{m}_t^k + b_z)
\end{equation}
\begin{equation}\label{eq:gating2}
i_t^k = \sigma (W_i \hat{m}_t^k + b_i - 1)
\end{equation}
\begin{equation}\label{eq:gating3}
f_t^k = \sigma (W_f \hat{m}_t^k + b_f + 1)
\end{equation}
\begin{equation}\label{eq:gating4}
m_{t+1}^k = m_{t}^k \odot f_t^k + z_t^k \odot i_t^k
\end{equation}

Finally, the new memory state $M^{t+1}$ is projected with a feedforward layer ($\mathrm{FFN}$). We replace the standard residual connection with the gating mechanism explained above. Note that the parameters are not tied, so this gating mechanism is not attached to the previous one.




\paragraph{Output Decoding:}

\begin{figure}
  \centering
  \includegraphics[width=0.20\textwidth]{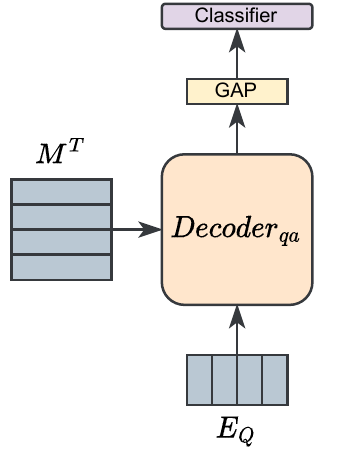}
  \vspace{-0.25cm}
  \caption{The output decoder takes the question $Q$ and the last memory state $M^T$ as inputs to obtain an answer $A$ using a classifier.}
  \label{fig:decoder-qa}
\end{figure}

After the model has processed the entire input stream and obtained our memory fed $M^T$, we use a decoder to obtain the answer (Figure~\ref{fig:decoder-qa}). We use a standard transformer decoder \cite{NIPS2017_3f5ee243}, which allows contextualizing the question token embeddings and aggregating question-relevant clues from memory (Eq.~\ref{eq:decoder}). Then, we perform a global average pooling (GAP) of the resulting representations $H_Q$ and use a classifier $W_{clf}$ to predict the answer (Eq.~\ref{eq:clf}) by optimizing a loss $\mathcal{L}_{qa}$. To allow $H_Q$ to capture enough information to answer the question, we perform multi-hop reasoning by iteratively updating $H_Q$ states with the same decoder. This is equivalent to having a multilayer decoder with tied weights.

\begin{equation}\label{eq:decoder}
H_Q = Decoder_{qa}(E_Q,M^{T})
\end{equation}
\begin{equation}\label{eq:clf}
A = W_{clf}(H_Q)
\end{equation}

\subsection{Rehearsal and Anticipation Mechanisms}

Rehearsal and anticipation are important processes that occur in our brain to refresh memory and prevent forgetting. On the one hand, rehearsal consists of continually bringing to our mind information already experienced to strengthen those memories \cite{Waugh1965,Craik1973}.
On the other hand, anticipation is like imagining the future. It has been found that anticipation can fire memory-forming regions of the brain — even before an event has occurred \cite{Mackiewicz2006,Wittmann2007}. 
Furthermore, rehearsal and anticipation were found to be related, might be using the same machinery \cite{Cole2015}, and guided by coreference information \cite{jaffe-etal-2020-coreference}.

Motivated by these findings, we propose implementing rehearsal as the reconstruction of the past \cite{pmlr-v139-zhang21ac} and anticipation as the prediction of the future \cite{oord2018representation,araujo-etal-2021-augmenting}. To use the same machinery, we pose these processes as masked modeling tasks that predict past and future coreference-related tokens.

\begin{figure}
  \centering
  \includegraphics[width=0.23\textwidth]{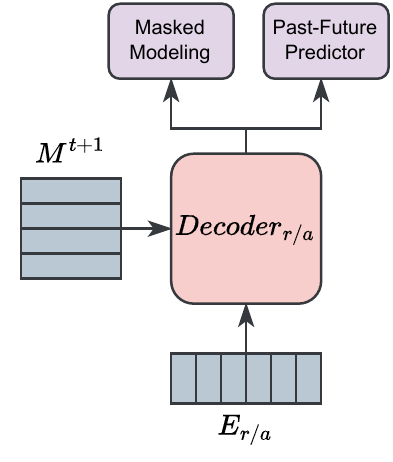}
  \vspace{-0.25cm}
  \caption{Our rehearsal and anticipation (r/a) decoder receives a masked input $E_{r/a}$ and the more recent memory updated $M^{t+1}$ to predict the masked tokens and whether the segment belongs to the past or future.}
  \label{fig:decoder-ra}
\end{figure}

\paragraph{Rehearsal:} This process is performed at step $t$ by randomly selecting a previous ($< t$) segment to mask some of its tokens. Using a standard transformer decoder (Figure~\ref{fig:decoder-ra}), we compute an input segment representation $E_r$ to obtain a contextual representation $H_r$. This process aggregates relevant clues through cross-attention from the updated memory $M^{t+1}$ to the representations (Eq.~\ref{eq:decoder-ra}). 

\paragraph{Anticipation:} This process is performed similarly to rehearsal (Eq.~\ref{eq:decoder-ra}), but the next ($t+1$) segment embedding $E_a$ is used instead. We use the same decoder to compute the representation $H_a$. 

\begin{equation}\label{eq:decoder-ra}
H_{r/a} = Decoder_{r/a}(E_{r/a},M^{t})
\end{equation}

where $r$ and $a$ denote rehearsal and anticipation, respectively.
The final hidden vectors ($H_r$ or $H_a$) corresponding to masked tokens are fed into a classifier to predict the actual token. 

\paragraph{What to rehearse and anticipate:} Our model has to store relevant information to predict future elements and, at the same time, avoid forgetting crucial information to predict the elements of previous segments.
For this to happen, it is important to mask significant tokens. Unlike the standard masked modeling, we do not mask tokens randomly but instead focus on masking conference-related tokens (for instance, person, object, or event).

The idea stems from the suggestion that coreference information may be helpful in retrieving probable antecedents from memory and in informing expectations about future words in language comprehension \cite{jaffe-etal-2020-coreference}.
Intuitively, by focusing on coreference relationships, our model can anticipate a coreferent and link it to the past through memory and rehearsal, also connecting the related pieces of discourse information.

\paragraph{Separating the tasks:} Since the task is the same for both mechanisms, the model would learn to predict the mask regardless of whether it is past or future information. Therefore, we include an additional signal to teach the model to distinguish where the segment comes from. We use the [CLS] token to create a binarized prediction task to classify whether the segment belongs to the past or the future.
As a result, we obtain three loss functions that we additively combine to obtain our final self-supervised modeling loss $\mathcal{L}_{ssm}$ (Eq.~\ref{eq:loss-ssm}). Our loss function is used in conjunction with the QA loss $\mathcal{L}_{qa}$ to optimize the whole model (Eq.~\ref{eq:loss}).

\begin{equation}\label{eq:loss-ssm}
\mathcal{L}_{ssm} = \mathcal{L}_{anticipation} + \mathcal{L}_{rehearsal} + \mathcal{L}_{binary}
\end{equation}
\begin{equation}\label{eq:loss}
\mathcal{L} = \mathcal{L}_{qa} + \mathcal{L}_{ssm}
\end{equation}




\paragraph{Novelty of our proposed method:} Although our model builds on the RM setup and model \cite{pmlr-v139-zhang21ac} incorporating anticipation along with rehearsal, we introduce important novelties. 

RM trains an additional QA model for each downstream task (\textit{history sampler}), using the complete input text to determine the important items to answer a question and then use them to rehearse. 
A key novelty of our method is that the memory uses coreference-related tokens that can be obtained by general-purpose part-of-speech (POS) taggers, making it task agnostic as we do not need to train additional models. By rehearsing and anticipating this information, we induce the model to softly learn coreference relationships, which is important for discourse comprehension.

In addition, RM updates its memory using single-head cross-attention and a disaggregated GRU network. Instead, our model implements self-attention, cross-attention, and gating in the same memory module, providing better integration of the incoming information with the one in the memory. 
Finally, our model is light and efficient due to the above features. Our memory module is parallelizable as it does not use a sequential GRU model and does not need to train additional QA models to decide what to rehearse or anticipate.

\section{Experiments}

\subsection{Experimental Settings}

\paragraph{Implementation Details:}

We use PyTorch to implement our model. 
We set the head number in all attention layers to 8, our decoders have 3 weight-sharing layers and K=20 memory slots. We tie the input embedding and output embedding as it has been demonstrated that it helps to improve language models \cite{press-wolf-2017-using} and memory networks \cite{liu-etal-2017-improving}. Also, this avoids having multiple output embeddings for the masked and QA modeling, reducing the model size.

We initialize the weights and bias of the gating mechanism using a truncated normal distribution with a mean of 0 and a standard deviation of 0.1, and add a constant of -1 and +1 to the input (Eq.~\ref{eq:gating2}) and forget gates (Eq.~\ref{eq:gating3}).
We use Adam optimizer with a learning rate of 4e-4 to train our model for 300 epochs with a batch size of 128. RAM for short sequence QA has $d_{model}=128$, and for long sequence QA has $d_{model}=256$, resulting in 1.2M and 3M parameters, respectively.

For the anticipation and rehearsal masked modeling, we use a POS tagging model to compute coreference-related tokens (nouns, pronouns, and verbs) for all the datasets. We rely on the FLAIR library \cite{akbik-etal-2019-flair}, which provides fast and accurate POS tagging models.
We mask coreference-related tokens up to a maximum of 40\% of segment tokens, which have been shown to be beneficial \cite{wettig-etal-2023-mask}.

\paragraph{Baselines:}
We compare our model with very well-known and recent memory networks that incrementally process the input data: DNC \cite{Graves2016}, NUTM \cite{le2018learning}, H-Mem \cite{10.5555/3495724.3497539}, DAM \cite{PARK202133}, STM \cite{pmlr-v119-le20b}, CT \cite{Rae2020Compressive}, and RM \cite{pmlr-v139-zhang21ac}. 
We borrowed the setup of \citet{pmlr-v139-zhang21ac}, in which baselines and memory size were adapted for a fair comparison with our model.
Regarding the specific hyperparameters, the NUTM core number was set to 4, the STM query number was set to 8, and the DAM memory block number was set to 2. The CT model uses a 3-layer transformer and a compression ratio of 5. Besides, all the models were set to K=20.

We also include direct methods that solve the task by accessing the entire input. We specify them in the section of each task explored.

\begin{table}
\centering
\begin{tabular}{c|c|c}
\toprule
Method & Mean Error & Best Error \\
\midrule
DNC$^\dagger$    & 16.70 $\pm$ 7.60       & 3.8        \\
NUTM$^\dagger$   & 5.60 $\pm$ 1.90        & 3.3        \\
H-Mem  & 8.93 $\pm$ 0.73      & 7.65       \\
DAM$^\dagger$ & 1.53 $\pm$ 1.33      & 0.16       \\
STM$^\dagger$    & 0.39 $\pm$ 0.18      & 0.15       \\
CT$^\dagger$     & 0.81 $\pm$ 0.26      & 0.34       \\
RM$^\dagger$     & 0.33 $\pm$ 0.15      & 0.12       \\
RAM    & \textbf{0.25 $\pm$ 0.16}     & \textbf{0.10 }      \\
\bottomrule
\end{tabular}
\caption{
Test error rates (in \%) on the 20 bAbI QA tasks for models using 10k training examples. We report mean $\pm$ std. and best error over 10 runs. $^\dagger$~is reported from \citet{pmlr-v139-zhang21ac}.
}
\label{tab:babi}
\end{table}

\subsection{Short Sequence QA}

For short sequence QA, we use the bAbI dataset \cite{arxiv.1502.05698}, a synthetic benchmark widely used to evaluate memory networks. This dataset consists of 20 short-sequence reasoning tasks (less than 100 words) that have to be solved with a common architecture. We use the percent error rate as the metric, which would be the complement of accuracy.
Table~\ref{tab:babi} shows the results. Our model achieves the best result compared to all baselines over 10 runs. Also, RAM has a low variance, being comparable with the most competitive models.

\subsection{Long Sequence QA}

For long sequence QA, we use NarrativeQA \cite{kocisky-etal-2018-narrativeqa}, a dataset with long input contents. It contains about 1,5000 stories and corresponding summaries (more than 600 words). In addition, it includes around 47,000 questions. We adopt the multiple-choice form proposed by the authors. Given a question associated with a summary, the idea is to retrieve the correct answer from a pool of answer candidates drawn from the associated questions.
We use mean reciprocal rank (MRR) \cite{voorhees-tice-2000-trec} as the metric to measure how far down the ranking the first relevant answer is.
We include two direct models: AS Reader \cite{kadlec-etal-2016-text} and E2E-MN \cite{NIPS2015_8fb21ee7} as additional baselines.

Table~\ref{tab:narrative} shows the results for the validation and test sets. Our model achieves 8.46\% and 7.38\% percentage differences of improvement with respect to the memory baseline RM for validation and test set, respectively. Note that our model also outperforms direct QA methods, being 9.49\% and 7.73\%, the percentage difference for validation and test set. These results demonstrate that incremental memory processing constitutes an effective and efficient approach over direct methods when well-designed.

\begin{table}
\centering
\begin{tabular}{c|c|c|c}
\toprule
\textbf{Method} & \textbf{Setting} & \textbf{Val} & \textbf{Test} \\
\midrule
AS Reader$^\dagger$       & Direct           & 26.9             & 25.9              \\
E2E-MN$^\dagger$          & Direct           & 29.1             & 28.6              \\
\midrule
DNC$^\dagger$             & Memory           & 25.8             & 25.2              \\
NUTM$^\dagger$            & Memory           & 27.7             & 27.2              \\
HMem                      & Memory           & 26.2             & 25.5              \\
DAM$^\dagger$             & Memory           & 28.1             & 27.5              \\
STM$^\dagger$             & Memory           & 27.2             & 26.7              \\
CT$^\dagger$              & Memory           & 28.7             & 28.3              \\
RM$^\dagger$              & Memory           & 29.4             & 28.7              \\
RAM                       & Memory           & \textbf{32.0}             & \textbf{30.9}              \\
\bottomrule
\end{tabular}
\caption{
Mean reciprocal rank (in \%) on Narrative QA for all the models. $^\dagger$~is reported from \citet{pmlr-v139-zhang21ac}.
}
\label{tab:narrative}
\end{table}

\begin{table}
\centering
\begin{tabular}{c|c|c}
\toprule
\textbf{Method} & \textbf{Setting} & \textbf{Test} \\
\midrule
E-MN$^\dagger$            & Direct           & 27.9 \\
E-SA$^\dagger$            & Direct           & 31.8 \\
HCRN$^\dagger$            & Direct           & 37.6 \\
\midrule
DNC$^\dagger$             & Memory           & 30.3 \\
NUTM$^\dagger$            & Memory           & 33.1 \\
HMem                      & Memory           & 31.9 \\
DAM$^\dagger$             & Memory           & 32.4 \\
STM$^\dagger$             & Memory           & 33.7 \\
CT$^\dagger$              & Memory           & 35.4 \\
RM$^\dagger$              & Memory           & 36.3 \\
RAM                       & Memory           & \textbf{37.4} \\
\bottomrule
\end{tabular}
\caption{
Accuracy (in \%) on ActivityNet-QA for all the models. $^\dagger$~is reported from \citet{pmlr-v139-zhang21ac}.
}
\label{tab:vqa}
\vspace{-2mm}          
\end{table}

\begin{table*}[htbp]
\centering
\begin{tabular}{c|c|c|c}
\toprule
Method & bAbI (Error Rate) & NarrativeQA (MRR) & ActivityNet-QA (Acc) \\
\midrule
RAM                                   & 0.25 $\pm$ 0.16 & 30.9  & 37.4 \\
RAM (random masking)                  & 0.28 $\pm$ 0.19 & 30.6  & 37.4 \\
RAM w/o $\mathcal{L}_{anticipation}$  & 0.31 $\pm$ 0.17 & 28.9  & 35.9 \\
RAM w/o $\mathcal{L}_{rehearsal} $    & 0.34 $\pm$ 0.17 & 29.7  & 36.4 \\
RAM w/o $\mathcal{L}_{ssm}$           & 0.37 $\pm$ 0.18 & 28.0  & 35.0 \\
\bottomrule
\end{tabular}
\caption{
Ablation results for all the datasets. All results are obtained on test sets.
}
\label{tab:ablation}
\vspace{-2mm}          
\end{table*}

\subsection{Video QA}

For video QA, we use ActivityNet-QA \cite{Yu2019}, which consists of 58,000 QA pairs about 5,800 complex web videos, derived from the popular ActivityNet dataset. We used the same procedure as for the bAbI task to predict the answer. 
As evaluation metric for the test set, we use accuracy.

As the input stream is video, we adapt our network to make it work with the visual modality. Following the original setup of ActivityNet-QA \cite{Yu2019}, we use fixed sampling to obtain 20 frames 
as the sequence that our model will process incrementally.
To encode the images and decode the masked tokens, we closely follow the framework proposed by \citet{Xie_2022_CVPR} for masked image modeling. The images of the stream are linearly projected to obtain a sequence of patch embeddings with a 32 × 32 pixel resolution, to which position embeddings are added. 

For the rehearsal and anticipation tasks, we use a learnable mask token vector to replace each masked patch. To decode the masked patches, we use a linear layer to yield patch logits to compute the L1 loss considering only the masked patches.
The inputs are images, so we no longer have content words to mask. As an alternative, we propose using the objects in the scene as tokens to mask. Therefore, we use the object detector YOLO \cite{glenn_jocher_2022_7347926} to compute the patches of the objects in a frame. We use all the object categories but constrain the masking up to 40\% of tokens.

Table~\ref{tab:vqa} shows the test set results. Our model achieves a percentage difference improvement of 2.98\% with respect to the model RM and 5.49\% with respect to the model CT. Note that our model performs almost comparable to the best direct baseline, HCRN, having only a 0.5\% percentage difference of improvement.

\section{Further Experimentation}

\subsection{Ablation Study}
The ablation study allows us to understand which components or configurations impact our model's performance most. We are interested in analyzing the impact of the proposed pretext tasks and masking strategy.

Table~\ref{tab:ablation} shows the four ablation results we performed compared with the default model. First, we found that randomly masking tokens in the masked modeling task slightly affects the results for both bAbI and NarrativeQA. However, for ActivityNet-QA, the impact is negligible, suggesting that random masking is good enough to capture relevant visual information, as shown in \cite{Xie_2022_CVPR}.

Also, we found that when we remove the pretext tasks entirely (w/o $\mathcal{L}_{ssm}$), the performance drops dramatically, resulting in being competitive with the no-self-supervised memory STM. By removing only the anticipation task (w/o $\mathcal{L}_{anticipation}$) or only the rehearsal task (w/o $\mathcal{L}_{rehearsal}$), the performance approaches the RM model. This means that the backbone of our model is more robust, possibly due to the gating mechanism or the fusion module. Note that removing the anticipation task impacts performance more than removing the rehearsal task, but both mechanisms help to boost performance. This supports that rehearsal is a crucial mechanism for memory networks \cite{PARK202133,pmlr-v139-zhang21ac} and confirms that anticipation is a helpful signal to improve memorization.

\begin{figure}
    \centering
\vspace{-2mm}        
    \subfigure[NarrativeQA]{\includegraphics[width=0.23\textwidth]{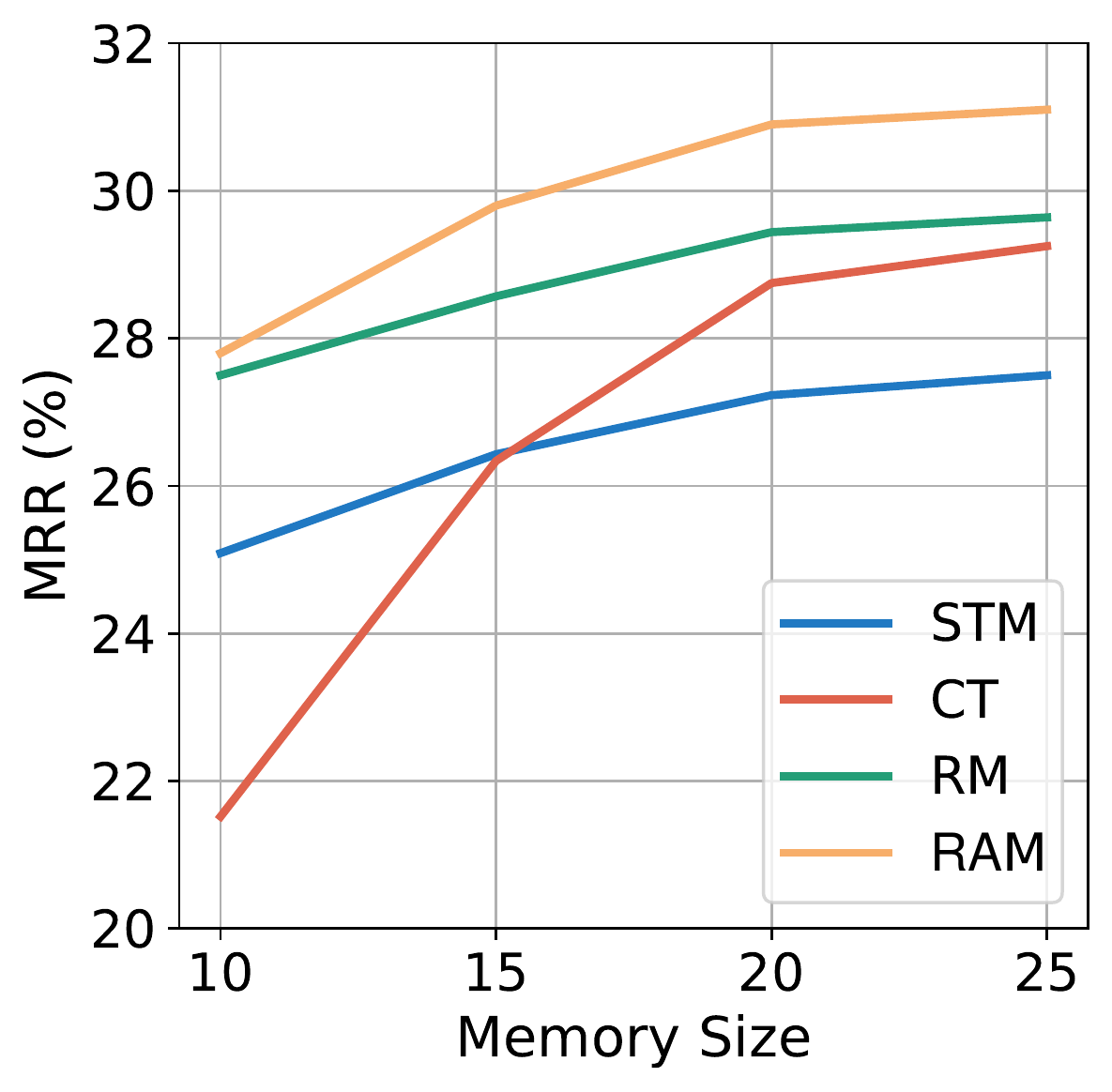}    \label{fig:1}} 
    \subfigure[ActivityNet-QA]{\includegraphics[width=0.23\textwidth]{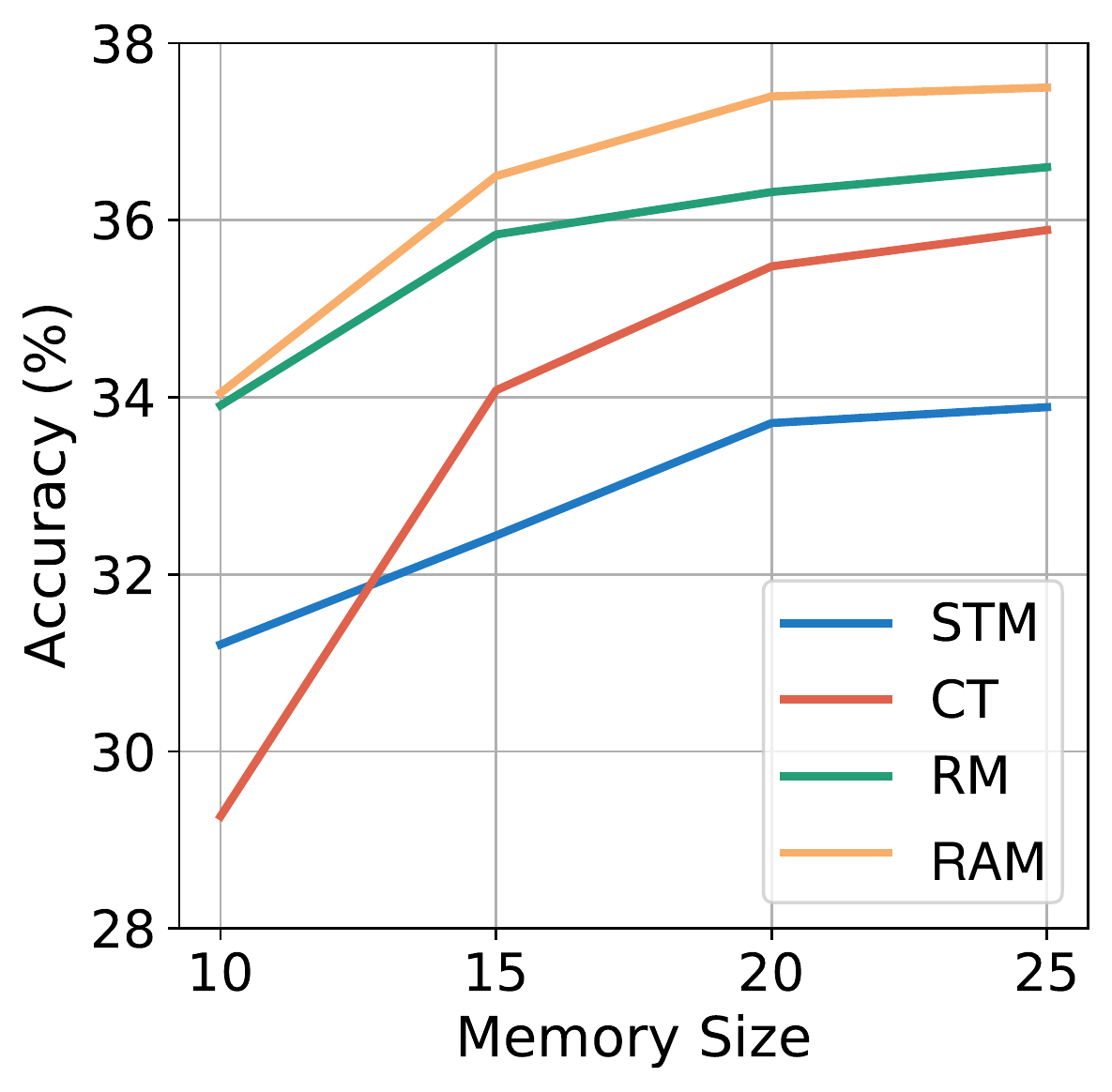}
    \label{fig:2}} 
    \caption{Performance of models across different memory sizes for (a) NarrativeQA and (b) ActivityNet-QA.}
    \label{fig:memsize}
\vspace{-4mm}    
\end{figure}

\subsection{Effect of Memory Size}

Figure~\ref{fig:memsize} shows the results of our model and some of the more robust memory-based baselines for NarrativeQA and ActivityNet-QA. We evaluate all the models using 10 to 25 memory slots. We do not include the results of the bAbI task because we found that the performance does not vary substantially when changing the memory size of the models.

We found that the performance for all models gradually increases with growing memory size.
As shown, the CT model is the most affected when the size is reduced, followed by the STM model.
The behavior of STM, RM, and RAM is slightly similar but with a difference in performance. We found that our model with low memory is almost equivalent to the performance of RM, but as memory capacity increases, the performance gap also increases. We noticed that memory slots higher than 25 do not show substantial improvements, suggesting that 20-25 slots are adequate for these tasks.

\subsection{What is the Attention Capturing?}

This section aims to reveal what kind of interactions are taking place between the inputs and the memory slots by visualizing the attention weights.
Our approach uses cross-attention to both integrate information in the memory and also to retrieve query-relevant information from the memory to obtain the answer.
Therefore, we analyze the attention of those layers obtained from random examples.

\begin{figure}
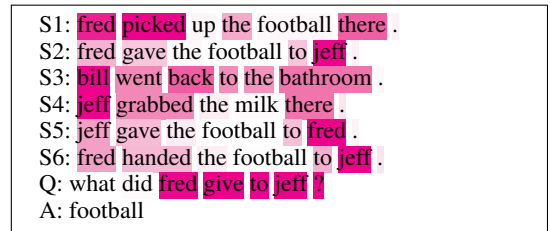
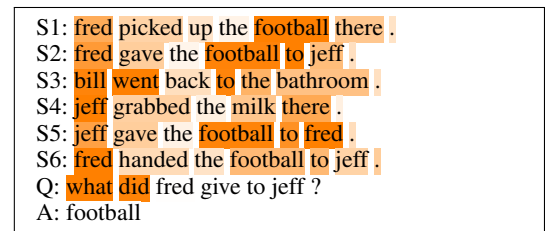

\centering 

\subfigure[Slot 8]{
\fbox{
\begin{small}
\setlength{\fboxsep}{0pt}\colorbox{white!0}{
\parbox{0.40\textwidth}{

\colorbox{magenta!0.0}{\strut S1:} \colorbox{magenta!94.69063777870217}{\strut fred} \colorbox{magenta!100.0}{\strut picked} \colorbox{magenta!4.524483417796941}{\strut up} \colorbox{magenta!41.72170734203257}{\strut the} \colorbox{magenta!3.4782473523942037}{\strut football} \colorbox{magenta!77.45252495283641}{\strut there} \colorbox{magenta!5.390245628923596}{\strut .} 

\colorbox{magenta!0.0}{\strut S2:} \colorbox{magenta!37.10924857604647}{\strut fred} \colorbox{magenta!27.508923965596594}{\strut gave} \colorbox{magenta!3.4312967724537837}{\strut the} \colorbox{magenta!0.7491060085240506}{\strut football} \colorbox{magenta!28.580828353718406}{\strut to} \colorbox{magenta!100.0}{\strut jeff} \colorbox{magenta!7.097033970980866}{\strut .} 

\colorbox{magenta!0.0}{\strut S3:} \colorbox{magenta!100.0}{\strut bill} \colorbox{magenta!59.249472074680895}{\strut went} \colorbox{magenta!78.52264241911482}{\strut back} \colorbox{magenta!56.500969920517576}{\strut to} \colorbox{magenta!53.16889326280189}{\strut the} \colorbox{magenta!68.30078509949912}{\strut bathroom} \colorbox{magenta!3.4285455836654632}{\strut .} 

\colorbox{magenta!0.0}{\strut S4:} \colorbox{magenta!100.0}{\strut jeff} \colorbox{magenta!59.11544654180556}{\strut grabbed} \colorbox{magenta!7.39050363525611}{\strut the} \colorbox{magenta!2.039232042760828}{\strut milk} \colorbox{magenta!68.76428334117564}{\strut there} \colorbox{magenta!12.15396539353332}{\strut .} 

\colorbox{magenta!0.0}{\strut S5:} \colorbox{magenta!47.23826268229373}{\strut jeff} \colorbox{magenta!32.813452717159194}{\strut gave} \colorbox{magenta!8.524789856769235}{\strut the} \colorbox{magenta!0.4445961820778912}{\strut football} \colorbox{magenta!30.642389055822818}{\strut to} \colorbox{magenta!100.0}{\strut fred} \colorbox{magenta!5.615080336816695}{\strut .} 

\colorbox{magenta!0.0}{\strut S6:} \colorbox{magenta!46.1079971719564}{\strut fred} \colorbox{magenta!34.28156511575179}{\strut handed} \colorbox{magenta!8.001109717413515}{\strut the} \colorbox{magenta!0.7048550076985899}{\strut football} \colorbox{magenta!38.31586555923174}{\strut to} \colorbox{magenta!100.0}{\strut jeff} \colorbox{magenta!12.00984404340324}{\strut .} 

\colorbox{magenta!0.0}{\strut Q:} \colorbox{magenta!0.0}{\strut what} \colorbox{magenta!2.381382459947762}{\strut did} \colorbox{magenta!99.19653982254853}{\strut fred} \colorbox{magenta!100.0}{\strut give} \colorbox{magenta!100.0}{\strut to} \colorbox{magenta!99.90144767848223}{\strut jeff} \colorbox{magenta!99.99852262233895}{\strut ?} 

\colorbox{magenta!0.0}{\strut A:} \colorbox{magenta!0.0}{\strut football} 

}}
\end{small}
}
\label{fig:slot8}}

\subfigure[Slot 18]{
\fbox{
\begin{small}
\setlength{\fboxsep}{0pt}\colorbox{white!0}{\parbox{0.40\textwidth}{
\colorbox{orange!0.0}{\strut S1:} \colorbox{orange!65.07691220058845}{\strut fred} \colorbox{orange!34.52255801310664}{\strut picked} \colorbox{orange!20.87745938128921}{\strut up} \colorbox{orange!9.26371220969107}{\strut the} \colorbox{orange!100.0}{\strut football} \colorbox{orange!43.83769791710839}{\strut there} \colorbox{orange!22.345374163427362}{\strut .} 

\colorbox{orange!0.0}{\strut S2:} \colorbox{orange!78.95159783419939}{\strut fred} \colorbox{orange!38.81614268800273}{\strut gave} \colorbox{orange!7.852743767878774}{\strut the} \colorbox{orange!96.98583507258158}{\strut football} \colorbox{orange!100.0}{\strut to} \colorbox{orange!52.13462395534596}{\strut jeff} \colorbox{orange!25.293147987934496}{\strut .} 

\colorbox{orange!0.0}{\strut S3:} \colorbox{orange!96.17707785450331}{\strut bill} \colorbox{orange!100.0}{\strut went} \colorbox{orange!25.449627723969176}{\strut back} \colorbox{orange!95.52299295350795}{\strut to} \colorbox{orange!59.91824995439319}{\strut the} \colorbox{orange!48.30965264684231}{\strut bathroom} \colorbox{orange!24.282725008115243}{\strut .} 

\colorbox{orange!0.0}{\strut S4:} \colorbox{orange!100.0}{\strut jeff} \colorbox{orange!36.78649463948366}{\strut grabbed} \colorbox{orange!12.099935794649424}{\strut the} \colorbox{orange!44.75364833272684}{\strut milk} \colorbox{orange!67.6336313772307}{\strut there} \colorbox{orange!11.638595543025362}{\strut .} 

\colorbox{orange!0.0}{\strut S5:} \colorbox{orange!73.48021547135049}{\strut jeff} \colorbox{orange!38.69503426075373}{\strut gave} \colorbox{orange!10.325449502229857}{\strut the} \colorbox{orange!96.71926424298852}{\strut football} \colorbox{orange!86.46714056719055}{\strut to} \colorbox{orange!100.0}{\strut fred} \colorbox{orange!26.47257800006534}{\strut .} 

\colorbox{orange!0.0}{\strut S6:} \colorbox{orange!100.0}{\strut fred} \colorbox{orange!28.066672669877317}{\strut handed} \colorbox{orange!22.040712424807786}{\strut the} \colorbox{orange!53.739709400201995}{\strut football} \colorbox{orange!54.87879864032607}{\strut to} \colorbox{orange!34.75667148966031}{\strut jeff} \colorbox{orange!39.535777759010344}{\strut .} 

\colorbox{orange!0.0}{\strut Q:} \colorbox{orange!100.0}{\strut what} \colorbox{orange!97.61861419735673}{\strut did} \colorbox{orange!0.8034611501097919}{\strut fred} \colorbox{orange!0.0}{\strut give} \colorbox{orange!0.00000000007135638241374653}{\strut to} \colorbox{orange!0.09854924630832551}{\strut jeff} \colorbox{orange!0.0014845351827303727}{\strut ?} 

\colorbox{orange!0.0}{\strut A:} \colorbox{orange!0.0}{\strut football} 

}}
\end{small}

}
\label{fig:slot18}}

\caption{Attention visualization for a random example of bAbi task 5. Each subfigure shows the intensity of attention from a memory slot to each word of the sentence during reading and from each word of the question to the memory slot. Strongly highlighted tokens mean higher attention weights. The most relevant slots to answer the question are shown: Slot (a) 8 and (b) 18.}
\label{fig:attn-babi}
\vspace{-2mm}          
\end{figure}
\vspace{-2mm}

Figure~\ref{fig:attn-babi} shows an example from bAbI task 5. In this example, the objective is reasoning over argument relations. We found that 20 slots of memory capture several repeated patterns, so we analyze some of them. 
We notice that slot 8 is specialized to capture relationships between entities (e.g., fred, jeff), as more attention is seen on entity names and action verbs across all the steps. Besides, the attention from the query to slot 8 was related to the entities and the activity they carried out.

Regarding slot 18, we notice that this slot is specialized to capture the relation between entities and objects. We found that some slots became more specialized to capture specific items, such as this one for the "football" object. We can see that all entities have high attention and the word "football", which suggests that this slot is tracking the state of the football through the steps. Interestingly, we found that the question attention over the slots is very high for the word "what", showing that slot 8 contains information about the football.

\begin{figure}
\centering 

\subfigure[Slot 2]{
\fbox{
\begin{small}
\setlength{\fboxsep}{0pt}\colorbox{white!0}{\parbox{0.43\textwidth}{
\colorbox{cyan!0.0}{\strut S1:} 
\colorbox{cyan!43.9655172413793}{\strut Frankie} 
\colorbox{cyan!99.99852586206896}{\strut Ryan} 
\colorbox{cyan!5.29928922413793178}{\strut works} 
\colorbox{cyan!0.21928922413793178}{\strut as} 
\colorbox{cyan!6.0}{\strut a} 
\colorbox{cyan!43.9655172413793}{\strut page} 
\colorbox{cyan!80.0}{\strut boy} 
\colorbox{cyan!98.0}{\strut at} 
\colorbox{cyan!13.69476724137931}{\strut a} 
\colorbox{cyan!13.791629310344824}{\strut radio} 
\colorbox{cyan!13.78639655172414}{\strut station} 
\colorbox{cyan!24.70204784482759}{\strut located} 
\colorbox{cyan!3.21928922413793178}{\strut in} 
\colorbox{cyan!31.943427155172415}{\strut Hollywood} \colorbox{cyan!43.322737499999995}{\strut .} 

\colorbox{cyan!0.0}{\strut S2:}
\colorbox{cyan!17.35721645321753}{\strut His} 
\colorbox{cyan!59.82935306998455}{\strut friend} 
\colorbox{cyan!90.0}{\strut Jeff} 
\colorbox{cyan!7.21153118041846294}{\strut works} 
\colorbox{cyan!9.0}{\strut in} 
\colorbox{cyan!32.410099248322155}{\strut the} 
\colorbox{cyan!46.56795211580472}{\strut same} 
\colorbox{cyan!40.56795211580472}{\strut place} 
\colorbox{cyan!10.56795211580472}{\strut ,} 
\colorbox{cyan!4.89456618566033}{\strut but} 
\colorbox{cyan!5.988001455298398}{\strut as} 
\colorbox{cyan!13.298659556882402}{\strut a} 
\colorbox{cyan!75.05288279510462}{\strut porter} 
\colorbox{cyan!41.7900598552441}{\strut .} 

\colorbox{cyan!0.0}{\strut \vdots} 

\colorbox{cyan!0.0}{\strut S5:}
\colorbox{cyan!27.35721645321753}{\strut When} 
\colorbox{cyan!59.82935306998455}{\strut they} 
\colorbox{cyan!20.0}{\strut try} 
\colorbox{cyan!5.21153118041846294}{\strut to} 
\colorbox{cyan!8.3}{\strut help} 
\colorbox{cyan!20.410099248322155}{\strut the} 
\colorbox{cyan!46.56795211580472}{\strut station} 
\colorbox{cyan!46.56795211580472}{\strut receptionist} 
\colorbox{cyan!46.56795211580472}{\strut ,}
\colorbox{cyan!78.89456618566033}{\strut Anne} 
\colorbox{cyan!34.988001455298398}{\strut Mason}
\colorbox{cyan!4.988001455298398}{\strut ,} 
\colorbox{cyan!13.298659556882402}{\strut by} 
\colorbox{cyan!25.05288279510462}{\strut setting} 
\colorbox{cyan!37.35721645321753}{\strut up} 
\colorbox{cyan!69.82935306998455}{\strut a} 
\colorbox{cyan!10.0}{\strut false} 
\colorbox{cyan!1.21153118041846294}{\strut audition} 
\colorbox{cyan!4.0}{\strut for} 
\colorbox{cyan!42.410099248322155}{\strut the} 
\colorbox{cyan!26.56795211580472}{\strut position} 
\colorbox{cyan!46.56795211580472}{\strut as} 
\colorbox{cyan!49.89456618566033}{\strut singer} 
\colorbox{cyan!19.89456618566033}{\strut ,} 
\colorbox{cyan!14.988001455298398}{\strut they} 
\colorbox{cyan!13.298659556882402}{\strut are} 
\colorbox{cyan!55.05288279510462}{\strut almost} 
\colorbox{cyan!4.21153118041846294}{\strut fired} 
\colorbox{cyan!10.10576559020924}{\strut for} 
\colorbox{cyan!63.21153118041846294}{\strut their} 
\colorbox{cyan!50.10576559020924}{\strut antics} 

\colorbox{cyan!0.0}{\strut \vdots} 

\colorbox{cyan!0.0}{\strut Q:} 
\colorbox{cyan!67.35721645321753}{\strut What} 
\colorbox{cyan!0.0}{\strut happens} 
\colorbox{cyan!99.0}{\strut when} 
\colorbox{cyan!99.153118041846294}{\strut Frankie} 
\colorbox{cyan!0.0}{\strut and} 
\colorbox{cyan!0.0}{\strut Jeff} 
\colorbox{cyan!0.0}{\strut try} 
\colorbox{cyan!0.0}{\strut to} 
\colorbox{cyan!0.0}{\strut help} 
\colorbox{cyan!0.0}{\strut the} 
\colorbox{cyan!0.0}{\strut station} 
\colorbox{cyan!99.0}{\strut receptionist} 
\colorbox{cyan!99.0}{\strut ?} 

\colorbox{cyan!0.0}{\strut A:} \colorbox{cyan!0.0}{\strut They almost get fired .} 

}}
\end{small}
}
\label{fig:slot2}}

\subfigure[Slot 6]{
\fbox{
\begin{small}
\setlength{\fboxsep}{0pt}\colorbox{white!0}{\parbox{0.43\textwidth}{
\colorbox{pink!0.0}{\strut S1:} 
\colorbox{pink!33.9655172413793}{\strut Frankie} 
\colorbox{pink!49.99852586206896}{\strut Ryan} 
\colorbox{pink!55.29928922413793178}{\strut works} 
\colorbox{pink!0.21928922413793178}{\strut as} 
\colorbox{pink!6.0}{\strut a} 
\colorbox{pink!43.9655172413793}{\strut page} 
\colorbox{pink!5.0}{\strut boy} 
\colorbox{pink!3.0}{\strut at} 
\colorbox{pink!13.69476724137931}{\strut a} 
\colorbox{pink!10.791629310344824}{\strut radio} 
\colorbox{pink!15.78639655172414}{\strut station} 
\colorbox{pink!54.70204784482759}{\strut located} 
\colorbox{pink!13.21928922413793178}{\strut in} 
\colorbox{pink!10.943427155172415}{\strut Hollywood} \colorbox{pink!20.322737499999995}{\strut .} 

\colorbox{pink!0.0}{\strut S2:}
\colorbox{pink!37.35721645321753}{\strut His} 
\colorbox{pink!29.82935306998455}{\strut friend} 
\colorbox{pink!24.0}{\strut Jeff} 
\colorbox{pink!77.21153118041846294}{\strut works} 
\colorbox{pink!9.0}{\strut in} 
\colorbox{pink!32.410099248322155}{\strut the} 
\colorbox{pink!46.56795211580472}{\strut same} 
\colorbox{pink!40.56795211580472}{\strut place} 
\colorbox{pink!10.56795211580472}{\strut ,} 
\colorbox{pink!4.89456618566033}{\strut but} 
\colorbox{pink!15.988001455298398}{\strut as} 
\colorbox{pink!23.298659556882402}{\strut a} 
\colorbox{pink!15.05288279510462}{\strut porter} 
\colorbox{pink!71.7900598552441}{\strut .} 

\colorbox{pink!0.0}{\strut \vdots} 

\colorbox{pink!0.0}{\strut S5:}
\colorbox{pink!7.35721645321753}{\strut When} 
\colorbox{pink!29.82935306998455}{\strut they} 
\colorbox{pink!70.0}{\strut try} 
\colorbox{pink!35.21153118041846294}{\strut to} 
\colorbox{pink!80.3}{\strut help} 
\colorbox{pink!20.410099248322155}{\strut the} 
\colorbox{pink!16.56795211580472}{\strut station} 
\colorbox{pink!26.56795211580472}{\strut receptionist} 
\colorbox{pink!6.56795211580472}{\strut ,}
\colorbox{pink!58.89456618566033}{\strut Anne} 
\colorbox{pink!44.988001455298398}{\strut Mason}
\colorbox{pink!8.988001455298398}{\strut ,} 
\colorbox{pink!3.298659556882402}{\strut by} 
\colorbox{pink!75.05288279510462}{\strut setting} 
\colorbox{pink!67.35721645321753}{\strut up} 
\colorbox{pink!49.82935306998455}{\strut a} 
\colorbox{pink!10.0}{\strut false} 
\colorbox{pink!61.21153118041846294}{\strut audition} 
\colorbox{pink!14.0}{\strut for} 
\colorbox{pink!22.410099248322155}{\strut the} 
\colorbox{pink!16.56795211580472}{\strut position} 
\colorbox{pink!26.56795211580472}{\strut as} 
\colorbox{pink!89.89456618566033}{\strut singer} 
\colorbox{pink!29.89456618566033}{\strut ,} 
\colorbox{pink!4.988001455298398}{\strut they} 
\colorbox{pink!13.298659556882402}{\strut are} 
\colorbox{pink!5.05288279510462}{\strut almost} 
\colorbox{pink!4.21153118041846294}{\strut fired} 
\colorbox{pink!10.10576559020924}{\strut for} 
\colorbox{pink!33.21153118041846294}{\strut their} 
\colorbox{pink!40.10576559020924}{\strut antics} 

\colorbox{pink!0.0}{\strut \vdots} 

\colorbox{pink!0.0}{\strut Q:} 
\colorbox{pink!99.0}{\strut What} 
\colorbox{pink!99.0}{\strut happens} 
\colorbox{pink!0.0}{\strut when} 
\colorbox{pink!0.0}{\strut Frankie} 
\colorbox{pink!0.0}{\strut and} 
\colorbox{pink!0.0}{\strut Jeff} 
\colorbox{pink!99.0}{\strut try} 
\colorbox{pink!0.0}{\strut to} 
\colorbox{pink!99.0}{\strut help} 
\colorbox{pink!0.0}{\strut the} 
\colorbox{pink!0.0}{\strut station} 
\colorbox{pink!99.0}{\strut receptionist} 
\colorbox{pink!0.0}{\strut ?} 

\colorbox{pink!0.0}{\strut A:} \colorbox{pink!0.0}{\strut They almost get fired .} 

}}
\end{small}
}
\label{fig:slot6}}

\caption{Attention visualization for a random example of NarrativeQA. Each subfigure shows the intensity of attention from a memory slot to each word of the sentence during reading and from each word of the question to the memory slot. Strongly highlighted tokens mean higher attention weights. The most relevant slots to answer the question are shown: Slot (a)~2 and (b)~6. Vertical dots represent sentences that are not displayed.}
\label{fig:attn-nqa}
\vspace{-2mm}          
\end{figure}
\vspace{-2mm}        

Figure~\ref{fig:attn-nqa} shows an example from NarrativeQA. In this case, the example consists of several sentences, so we only include some relevant ones to solve the question. The example shows an outcome question inquiring about the specific result of a particular event. We notice that slot 2 mainly tracks people of the story, for instance, "Ryan", "Jeff" and "Anne". It is also possible to see that attention is paid to the pronoun tokens.

Regarding slot 6, we find that the attention is a bit more distributed across the tokens. However, more attention is paid to verbs (for example, "try" and "help"), which suggests that this slot specializes in events and participants. Note that attention to the question also presents the same behavior.

\begin{figure*}
\centering
\subfigure[Slot 7]{

\begin{minipage}{0.77\textwidth}
\includegraphics[width=12cm]{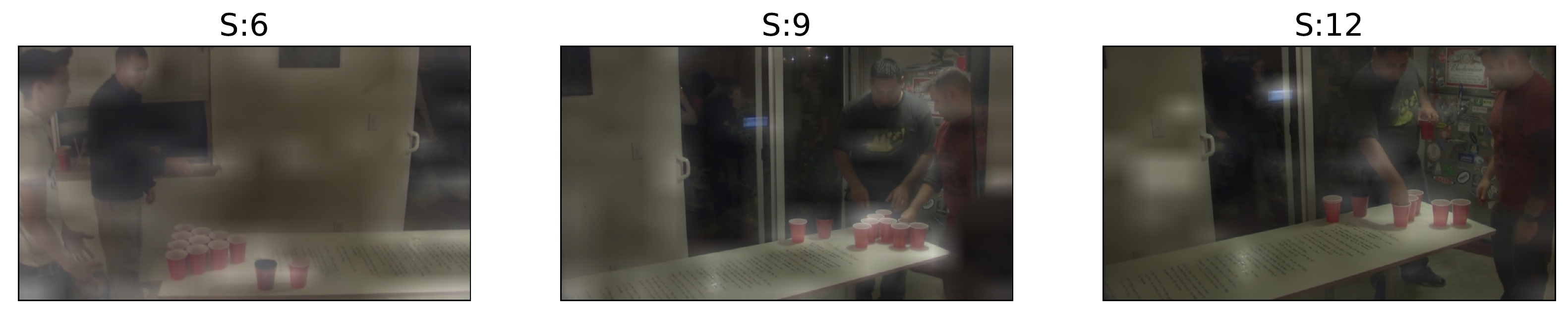}
\end{minipage}
\hfill
\begin{minipage}{0.37\textwidth}
\fbox{
\begin{footnotesize}

\setlength{\fboxsep}{0pt}\colorbox{white!0}{\parbox{0.45\textwidth}{
\colorbox{cyan!0.0}{\strut Q:} \colorbox{cyan!100.0}{\strut how} \colorbox{cyan!97.61861419735673}{\strut many} \colorbox{cyan!0.8034611501097919}{\strut people} \colorbox{cyan!0.0}{\strut are} \colorbox{cyan!0.00000000007135638241374653}{\strut playing} \colorbox{cyan!0.09854924630832551}{\strut games}
\colorbox{cyan!0.0014845351827303727}{\strut in} 
\colorbox{cyan!0.0014845351827303727}{\strut the}
\colorbox{cyan!0.0014845351827303727}{\strut video} 
\colorbox{cyan!0.0014845351827303727}{\strut ?} 

\colorbox{cyan!0.0}{\strut A:} \colorbox{cyan!0.0}{\strut 4} 

}}
\end{footnotesize}

}
\end{minipage}
\label{fig:vqa1}}

\subfigure[Slot 12]{

\begin{minipage}{0.77\textwidth}
\includegraphics[width=12cm]{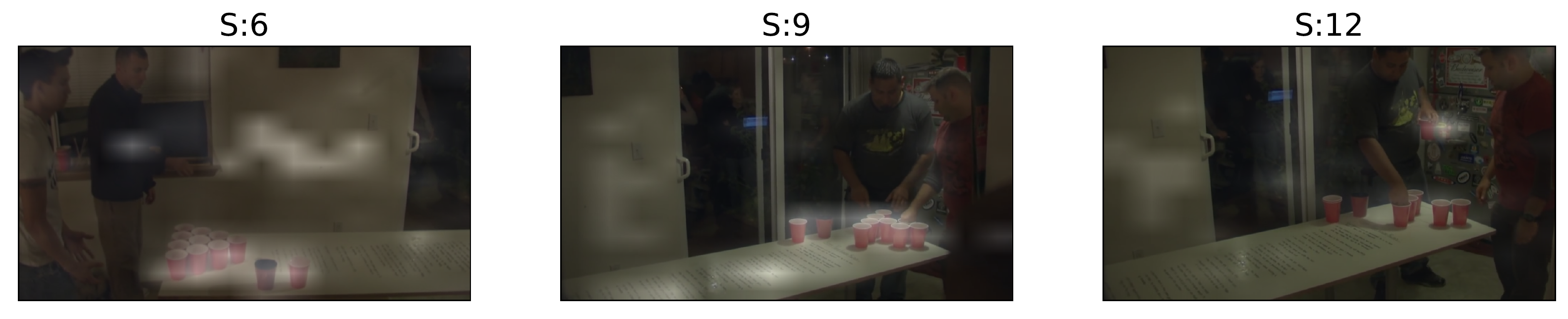}
\end{minipage}
\hfill
\begin{minipage}{0.37\textwidth}
\fbox{
\begin{footnotesize}

\setlength{\fboxsep}{0pt}\colorbox{white!0}{\parbox{0.45\textwidth}{
\colorbox{yellow!0.0}{\strut Q:} \colorbox{yellow!0.0}{\strut how} \colorbox{yellow!0.0}{\strut many} \colorbox{yellow!0.8034611501097919}{\strut people} \colorbox{yellow!!97.61861419735673}{\strut are} \colorbox{yellow!97.61861419735673}{\strut playing} \colorbox{yellow!0.09854924630832551}{\strut games}
\colorbox{yellow!0.0014845351827303727}{\strut in} 
\colorbox{yellow!0.0014845351827303727}{\strut the}
\colorbox{yellow!0.0014845351827303727}{\strut video} 
\colorbox{yellow!97.61861419735673}{\strut ?} 

\colorbox{yellow!0.0}{\strut A:} \colorbox{yellow!0.0}{\strut 4} 

}}
\end{footnotesize}

}
\end{minipage}
\label{fig:vqa2}} 

\caption{Attention visualization for a random example of ActivityNet-QA. Each subfigure shows the intensity of attention from a memory slot to each patch of the frame during reading and from each word of the question to the memory slot. Brighter regions of images mean higher attention weights. We show the most relevant slots and frames to answer the question: (a)~Slot 7 and (b)~Slot 12. }
\label{fig:attn-vqa}
\vspace{-4mm}          
\end{figure*}

Figure~\ref{fig:attn-vqa} shows an example from ActivityNet-QA. This example presents a counting task across the steps. In this case, the input stream is images. Our visualizations show bright regions when attention is high. We find a behavior similar to text-based models since some slots specialize in entities or the interaction between them or with objects.

We see that slot 7 is paying attention to the objects and the people around, suggesting that this slot captures the interaction between entities and objects across the steps. Note that slot 7 got the most attention for the words "how many", which supports the hypothesis that this slot tracks entity information.

As for slot 12, we find that it is specialized to capture information about the objects (cups, in this case). The focus is mainly on the cups; however, sometimes, in other places in the image. This suggests that attention tracks the interaction between objects in the scene.

These results support our initial hypothesis that slots could capture information about entities, objects, and relations between them. 
Also, it is possible to see that our model can track the state of those entities or objects across the steps, suggesting it can represent sequentially coreference-related information when updating the memory.

\vspace{-2mm}        


\section{Related Work}

\subsection{Question Answering}
In recent years there has been a significant advance in QA. This is thanks to the new datasets and novel deep learning architectures proposed to solve them.
One of the most important datasets is SQuAD \cite{rajpurkar-etal-2016-squad}, which proposes span prediction as the task for QA about a given paragraph.
SQuAD has been one of the reasons for progress in the field \cite{arxiv.2102.01065}, as several attention models have been proposed for this task. For example, BiDAF \cite{seo2017bidirectional} proposes a specialized bi-directional attention mechanism.

Recently, more complex QA tasks were proposed, such as long-range QA \cite{kocisky-etal-2018-narrativeqa,pang-etal-2022-quality} and commonsense QA \cite{talmor-etal-2019-commonsenseqa}. Also, other modalities were explored, using images \cite{Johnson_2017_CVPR,balanced_vqa_v2} or videos \cite{lei-etal-2018-tvqa,yu2019activityqa} as input. For these, large pre-trained models \cite{devlin-etal-2019-bert} have achieved outstanding results.

The models mentioned above always have access to the whole input. However, this is ineffective when the input is a data stream or a long sequence. For this reason, memory networks have been proposed to perform incremental processing to accumulate knowledge into memory and afterward perform the QA task \cite{miller-etal-2016-key,10.5555/3045390.3045643,han-etal-2019-episodic,pmlr-v139-zhang21ac}.

\subsection{Memory-augmented Neural Networks}

MANNs introduce models with external memory to store and access the past contents by differentiable operators. These models attempt to mimic human memory processes, bringing them closer to biological plausibility.
NTM \cite{arxiv.1410.5401} and DNC \cite{Graves2016} are well-known memorization and reasoning models which use memory instead of original input to generate inferences.

In this line of research, several papers have proposed new approaches for different NLP tasks, such as language modeling \cite{10.5555/3327757.3327832}, dialogue \cite{10.1145/3317612}, machine translation \cite{kaiser2017learning}, coreference resolution \cite{liu-etal-2019-referential}, question answering \cite{henaff2017tracking}, and visual question answering \cite{10.5555/3045390.3045643}. These approaches propose new ways of storing and retrieving information \cite{henaff2017tracking,Banino2020MEMO}, relational or associative mechanisms \cite{10.5555/3327757.3327832,10.5555/3495724.3497539,pmlr-v119-le20b}, and methods to improve memory representation and prevent forgetting \cite{PARK202133,pmlr-v139-zhang21ac}.

In recent years, some works have been proposed with the latter objective. DAM \cite{PARK202133} introduced a rehearsal task to ensure a current input is well stored by reproducing it using the memory. RM \cite{pmlr-v139-zhang21ac}  was proposed to tackle the long-term memorization problem, as this model can rehearse a current input and previous ones employing a recollection and familiarity task. Our work extends these approaches and builds on recent findings from neurocognitive science that posit that our memory is guided by rehearsal and anticipation of coreference information, and they might use the same machinery to improve human memorization \cite{Cole2015}. 
Unlike DAM and RM, our model proposes to perform not only rehearsal but also anticipation using masked modeling and a task to detect whether the information belongs to the future or the past. These mechanisms aim to improve memorization and prevent forgetting.

\section{Conclusion}

This article presents RAM, a QA memory model supported by two self-supervised pretext tasks to improve memory presentation. It uses masked modeling to mimic the human processes of rehearsal and anticipation of coreference information.
Our experiments on short sequence QA, long sequence QA, and video QA show that our model substantially outperforms previous memory models.
Our further experimentation demonstrates that the rehearsal and anticipation tasks help to improve memorization and that the model leverages coreference information to support the processes.

\section*{Limitations}

This work has some limitations regarding the architecture and the data used: Our model assumes that masked modeling could be used as rehearsal and anticipation tasks; however, other approaches could also be effective. We use transformer layers, so our model scalability is tied to the scalability of the transformer model. Also, our approach relies on obtaining additional annotation from pre-trained models for the masking process, so we are limited to the misprediction of those models. We only tested our model with the English language; further exploration of other languages would be valuable to validate the language-independent functionality of the model.

\section*{Ethics Statement}

Our article presents a novel approach and  has not been published in whole or in part elsewhere. 
The data used to train the model does not imply any violation of privacy.
The potential negative social impacts from this work are similar to any other NLP models.
Question answering models could potentially be used to create malicious chatbots.
This work does not include experimentation with humans or animals.

\section*{Acknowledgements}
This work was partially funded by European Research Council Advanced Grant 788506, FONDECYT grant 1221425, and National Center for Artificial Intelligence CENIA FB210017, Basal ANID.

\bibliography{anthology,custom}
\bibliographystyle{acl_natbib}




\end{document}